\documentclass{article}

\PassOptionsToPackage{numbers, compress}{natbib}

 \usepackage[main, final]{neurips_2026}

\usepackage[utf8]{inputenc} 
\usepackage[T1]{fontenc}    
\usepackage{graphicx}
\usepackage{booktabs}
\usepackage{enumitem}
\usepackage[accsupp]{axessibility}  
\usepackage{multirow}
\usepackage[table]{xcolor}  
\usepackage{pifont}
\definecolor{mred}{RGB}{238, 34, 12}
\definecolor{mgreen}{RGB}{1, 127, 0}
\definecolor{mblue}{RGB}{0, 77, 158}
\newcommand{\mredbf}[1]{\textcolor{mred}{\textbf{#1}}}
\newcommand{\mbluebf}[1]{\textcolor{mblue}{\textbf{#1}}}
\usepackage{amsmath}
\usepackage{arydshln} 
\usepackage{makecell}
\usepackage{titlesec}
\titlespacing*{\section}{0pt}{2pt plus 0.5pt minus 0.5pt}{2pt plus 0.5pt minus 0.5pt} 
\titlespacing*{\subsection}{0pt}{2pt plus 0.5pt minus 0.5pt}{2pt plus 0.5pt minus 0.5pt} 
\titlespacing*{\paragraph}
{0pt}    
{0.5em}    
{0.5em}    
\usepackage{hyperref}
\newcommand{\MYhref}[3][blue]{\href{#2}{\color{#1}{#3}}} 
\usepackage{orcidlink}
\usepackage{makecell}

\title{EditRefiner: A Human-Aligned Agentic Framework for Image Editing Refinement}

\author{
Zitong Xu\textsuperscript{1},
Huiyu Duan\textsuperscript{1†},
Yifei Nie\textsuperscript{3},
Mingda Du\textsuperscript{3},
Sijing Wu\textsuperscript{1},
Xiongkuo Min\textsuperscript{1},
\vspace{0.2em}
\\
Tianyi Zheng\textsuperscript{2},
Jian Zhang\textsuperscript{2},
Shusong Xu\textsuperscript{2},
Jinwei Chen\textsuperscript{2},
Bo Li\textsuperscript{2†},
Guangtao Zhai\textsuperscript{1†}
\vspace{0.5em}
\\
\textsuperscript{1}Shanghai Jiao Tong University
\vspace{0.2em}
\\
\textsuperscript{2}Vivo Mobile Communication Co., Ltd
\\
\vspace{0.2em}
\textsuperscript{3}University of Electronic Science and Technology of China
\\
\vspace{0.2em}
\textsuperscript{†}Corresponding Authors
}

\begin{document}
\maketitle
\begin{abstract}
Recent text-guided image editing (TIE) models have made remarkable progress, yet edited images still frequently suffer from fine-grained issues such as unnatural objects, lighting mismatch, and unexpected changes. Existing refinement approaches either rely on costly iterative regeneration or employ vision-language models (VLMs) with weak spatial grounding, often resulting in semantic drift and unreliable local corrections. To address these limitations, we first construct \textbf{EditFHF-15K}, a dataset of fine-grained human feedback for edited images, comprising (1) 15K images from 12 TIE models spanning 43 editing tasks, (2) 60K annotated artifact regions and 80K editing failure regions, each accompanied by textual reasoning, and (3) 45K mean opinion scores (MOSs) assessing perceptual quality, instruction following, and visual consistency. Based on EditFHF-15K, we propose \textbf{EditRefiner}, a hierarchical, interpretable, and human-aligned agentic framework that reformulates post-editing correction as a human-like perception-reasoning-action-evaluation loop. Specifically, we introduce: (1) a perception agent that detects contextual saliency maps of artifacts and editing failures, (2) a reasoning agent that interprets these perceptual cues to perform human-aligned diagnostic inference, (3) an action agent that uses the reasoning output to plan and execute localized re-editing, and (4) an evaluation agent that assesses the re-edited image and guides the action agent on whether further refinements are required. Extensive experiments demonstrate that EditRefiner consistently outperforms state-of-the-art methods in distortion localization, diagnose accuracy and human perception alignment, establishing a new paradigm for self-corrective and perceptually reliable image editing. The code is available at \MYhref[magenta]{https://github.com/IntMeGroup/EditRefiner}{https://github.com/IntMeGroup/EditRefiner}.
\end{abstract}

\section{Introduction}
The rapid advancement of text-guided image editing (TIE) allows flexible image modifications through natural language instructions \cite{nanobanana,seedream4,ACE, qwenedit, ip2p, Magicbrush}. Although some state-of-the-art editing models, such as Nano-Banana \cite{nanobanana}, Seedream4.0 \cite{seedream4}, Qwen-Image-Edit \cite{qwenedit}, \textit{etc.}, have achieved impressive performance, many edited images still suffer from subtle issues such as stiff limbs, lighting mismatch, unreadable text, and unintended changes. These flaws typically appear in localized areas of overall high-quality outputs, making them difficult to detect and costly to correct through full-image re-editing. Consequently, current TIE systems still lack an autonomous, fine-grained refinement framework, which remains a major barrier to real-world creative and industrial applications.

Recent research has explored three main strategies to improve the quality of TIE, including model fine-tuning \cite{ip2p,Magicbrush,HQ}, text prompt enhancement \cite{promptenhancer}, and reinforcement learning-based optimization \cite{editscore,editreward,xu2026edithf1mmillionscalerichhuman}. However, fine-tuning methods require large-scale, labor-intensive datasets; prompt enhancement approaches may introduce irrelevant or spurious information; and reinforcement learning techniques incur substantial computational cost. Moreover, although these approaches can improve overall realism, they lack explicit spatial reasoning and cannot reliably identify or correct local editing failures. With the emergence of vision-language models (VLMs) possessing advanced semantic reasoning, some works have explored using agentic framework to guide image generation \cite{liang2024rich,shen2026agentic,xu2025magicwanduniversalagentgeneration}. However, accurately assessing image editing is inherently more complex than generation alone, as it requires VLMs to compare the edited image with the source and interpret the editing instructions. Moreover, strong knowledge priors of VLMs can override visual evidence, causing hallucinated judgments, highlighting the need to align TIE refinement agent with human perception.

In this work, we introduce \textbf{EditFHF-15K}, a dataset with fine-grained human feedback on edited images, curated by trained annotators following a rigorous and standardized annotation protocol. EditFHF-15K contains \textit{15K} source images paired with edited images and their corresponding editing prompts, covering \textit{43 editing tasks} and \textit{13 state-of-the-art TIE models}. Based on the images, we further collect annotated \textit{60K artifact regions} and \textit{80K editing failure regions} for human-aligned perception agent development, with each region labeled using both semantic-level bounding boxes and pixel-level saliency maps. Each region is also accompanied by a human-annotated textual explanation to enable human-aligned reasoning agent construction. In addition, we collect \textit{45K MOSs} on perceptual quality, instruction following, and visual consistency for human-aligned evaluation agent establishment. Building upon EditFHF-15K, we propose \textbf{EditRefiner}, a hierarchical, interpretable, and human-aligned agentic framework that reformulates post-editing correction as a structured \textbf{perception-reasoning-action-evaluation loop}. 
Through iterative verification, EditRefiner fuses perceptual cues, semantic reasoning, controllable re-editing, and human-aligned evaluation into a coherent self-corrective process, enabling automatic correction of fine-grained flaws in image editing. We conduct experiments on EditFHF-15K and other public TIE benchmarks, refining edited images generated by state-of-the-art editing models. The consistent performance gains across extensive datasets validate the effectiveness and robustness of EditRefiner. 

The main contributions of this work include:
\begin{itemize}[left=12pt, labelsep=0.6em, labelwidth=0pt]
    \item We introduce EditFHF-15K, a dataset of fine-grained human feedback on edited images, comprising annotations of distortion regions, textual rationales, and preference ratings.
    \item We propose EditRefiner, a novel paradigm that reformulates post-editing as a perception-reasoning-action-evaluation loop, enabling automatic diagnosis and refinement for flaws.
    \item We design a collaborative four-agent system, where a perception agent performs context-aware flaw localization, a reasoning agent conducts human-aligned fine-grained diagnosis, an action agent executes controllable re-editing, and an evaluation agent assesses the refined results and guides further corrections.
    \item Extensive experiments on EditFHF-15K and other benchmarks demonstrate that EditRefiner consistently and effectively improves the quality of edited images, while performing regional cue extraction, semantic reasoning, and evaluation in better alignment with human preferences.
\end{itemize}

\section{Related work}
\subsection{Text-guided image editing}
Text-guided image editing has witnessed remarkable progress with the advent of large-scale diffusion models such as Stable Diffusion \cite{SD} and FLUX \cite{FLUX}. Early methods typically transform source image based on a target prompt, leveraging techniques such as attention control \cite{Masactrl,cds} and noise inversion \cite{pnp,OT,Flowedit}. Subsequent works adopt fine-tuning method \cite{ip2p,Magicbrush,HQ}, which allows models to modify style or semantic content directly from an instruction prompt. More recently, unified models \cite{qwenedit, omnigen2, dreamomni2, nanobanana, seedream4,wei2026skyworkunipic30unified} integrate understanding and generation within a single architecture, enabling flexible interpretation of editing instructions and producing highly realistic image edits. However, issues such as local artifacts, unnatural poses, and inconsistent lighting still occur \cite{lmm4edit,xu2026edithf1mmillionscalerichhuman}.

\subsection{Visual quality assessment}
Visual quality assessment (VQA) aims to evaluate visual content in alignment with human perception \cite{finevq,lmm4edit,harmonyiqa,wang2025lmm4lmm,wang2026quality}. In the context of AIGC content assessment, most existing approaches \cite{lmm4edit,wang2025lmm4lmm,xu2026edithf1mmillionscalerichhuman} rely on global quantitative metrics, lacking explicit localization and fine-grained evaluation of local defects. To address this limitation, recently works \cite{yang2025quality,liang2024rich,shen2026agentic} introduce predictors for local structural distortions. However, these methods are not designed for the image editing scenario, where evaluation must jointly consider the source image and the editing instruction.

\subsection{Multi-agent system}
Multi-agent systems have emerged with the advancement of VLMs for complex multimodal tasks \cite{llmagent2,agent1,agent2,agent3}. By enabling role specialization, collaboration, and iterative feedback, multi-agent frameworks can effectively decompose tasks, coordinate across agents, and progressively refine outputs \cite{zhu2024intelligent,li2025editthinkerunlockingiterativereasoning,xu2025magicwanduniversalagentgeneration,liu2025moa}. Such collaborative structures are particularly well-suited for scenarios requiring not only generation but also reasoning, assessing, and refining \cite{xu2025magicwanduniversalagentgeneration,ye2026agent}. In image generation applications, these capabilities are crucial for achieving high-quality results \cite{shen2026agentic}. However, human-aligned agent frameworks for image editing remain unexplored.

\begin{figure}
    \centering
    \includegraphics[width=1\linewidth]{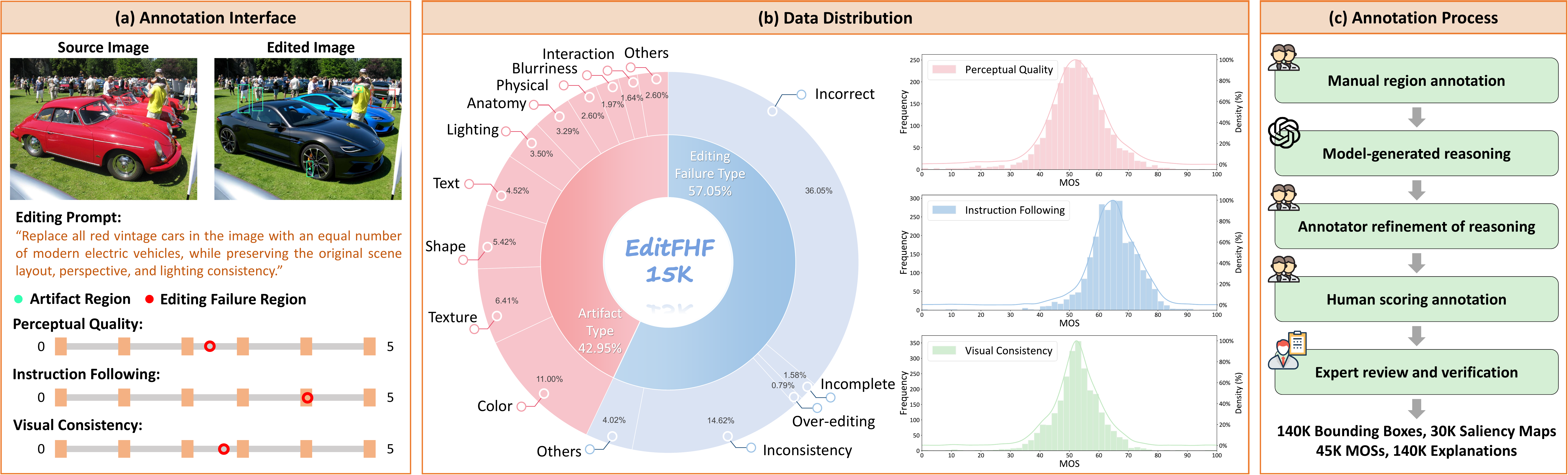}
    \caption{Overview of our EditFHF-15K. (a) An illustration of our annotation interface, (b) the distribution of flaw types and MOSs, (c) the human-AI collaborative annotation pipeline}
    \label{editfhf}
\end{figure}
\section{EditFHF-15K}
In this section, we introduce \textbf{EditFHF-15K}, a dataset of fine-grained human feedback on edited images. It comprises 15,287 edited images, each annotated with two saliency maps (artifact and editing failure) and three fine-grained scores (perceptual quality, instruction following, and visual consistency). The dataset also provides 60,318 bounding boxes for artifact regions and 80,345 bounding boxes for editing failure regions, with each box accompanied by a corresponding text describing the artifact or editing failure. This dataset facilitates the development of human-aligned systems capable of perception, reasoning, and evaluation of image editing.

\subsection{Image collection} 
Considering practical relevance and real-world usage, we select 43 image editing tasks covering diverse operations. We collect 18K source images with editing prompts, including 10K sampled from Pico-Banana-400K \cite{picobanana} and EBench-18K \cite{lmm4edit}, while the remaining 8K are gathered from publicly available photography websites, each accompanied by a textual description. Based on the image content and predefined editing tasks, we leverage the advanced VLM InternVL3.5 \cite{internvl3_5} to generate corresponding editing prompts, followed by careful manual verification and revision. 
We then select 13 advanced TIE models, including both open-source and closed-source methods with diverse backbone architectures, to generate edited images with varied characteristics. We then use LMM4Edit \cite{lmm4edit} to score the results and retain the top three images for each editing instance, filtering out low-quality outputs while ensuring instruction consistency. More details are provided in Section~\ref{collection} of the supplementary material.

\subsection{Human annotation}
\label{annotation}
To annotate the edited images, we first establish a rigorous annotation protocol. All annotators are required to undergo training and pass a qualification test. Only those who pass are allowed to participate in the annotation process. The overall annotation pipeline consists of three stages: regional annotation, textual annotation and scoring annotation.

\subsubsection{Regional annotation}

In the first stage, annotators are asked to identify artifact regions and editing failure regions, and annotate them using bounding boxes and disks. Specifically, bounding boxes are used to enclose the entire semantic object exhibiting a flaw, while disks with a radius of 1/20 of the image height are used to mark localized flawed areas. Each problematic semantic object is assigned one bounding box, and each bounding box must be associated with at least one disk.

The two types of regions are defined as follows: \textbf{(1) Artifact Region}: areas in the edited image that exhibit visual artifacts, such as unnatural textures, color inconsistencies, noise, or distortions. \textbf{(2) Editing Failure Region}: areas where the editing does not follow the instruction, including missing, incorrect, or unintended edits.

Each image is independently annotated by three annotators, who label the two region types separately. 
The results are then reviewed by an expert team. 
Finally, all disks are aggregated using Gaussian kernels to generate a pixel-level saliency map for each image. This stage produces both semantic-level bounding boxes and pixel-level saliency maps.

\subsubsection{Textual annotation}

In the second stage, we first employ the advanced VLM ChatGPT-5~\cite{gpt5} to generate initial descriptions for each flaw region. The model takes the source image, edited image, editing instruction, and bounding boxes as input. The bounding boxes serve as visual guidance, enabling more accurate and localized descriptions. Each prediction consists of a textual description that specifies the flaw type along with a corresponding reasoning statement, which explains the characteristics and visual details of the flaw within the associated region. These descriptions are then reviewed by three annotators in a round-robin manner, who may revise or rewrite them as needed. A description is finalized only when all three annotators reach agreement on its accuracy and completeness.

\subsubsection{Scoring annotation}

In the third stage, annotators are asked to rate each edited image across three evaluation dimensions: \textbf{(1) Perceptual Quality}: measures the overall visual quality of the edited image, including realism, absence of artifacts, structural integrity, color consistency, and richness of details. \textbf{(2) Instruction Following}: evaluates how well the edited image follows the given instruction, assessing whether the intended modifications are accurately and completely applied. \textbf{(3) Visual Consistency}: assesses how well the edited image maintains essential attributes of the source image beyond the intended edits, such as subject identity, key visual characteristics, and contextual consistency.

Each image is rated on a continuous 5-point scale by 15 annotators. During evaluation, annotators are provided with the source image, edited image, editing instruction, and the regional and textual annotations from previous stages, as shown in Figure~\ref{editfhf}(a). These annotations serve as references for scoring and are also implicitly validated during this process. Annotators can flag unreasonable annotations and any annotation identified as problematic by more than three annotators is discarded.

Following~\cite{subject}, ratings that deviate by more than two standard deviations from the mean are treated as outliers and removed. Annotators with more than 5\% outlier ratings are excluded. The remaining scores are normalized into Z-scores and linearly scaled to the range [0, 100]. The final score for each image is computed as:
\begin{equation}
    z_{ij} = \frac{s_{ij} - \mu_i}{\sigma_i}, \quad
    z_j = \frac{1}{N_j} \sum_{i=1}^{N_j} z_{ij}, \quad
    \text{Score}_j = \frac{100(z_j + 3)}{6}
\end{equation}
where $s_{ij}$ denotes the raw score assigned by the $i$-th annotator to the $j$-th image, $\mu_i$ and $\sigma_i$ are the mean and standard deviation of annotator $i$, and $N_j$ is the number of valid ratings for image $j$. This process yields reliable MOSs for each image across the three dimensions.


\begin{figure}
    \centering
    \includegraphics[width=1\linewidth]{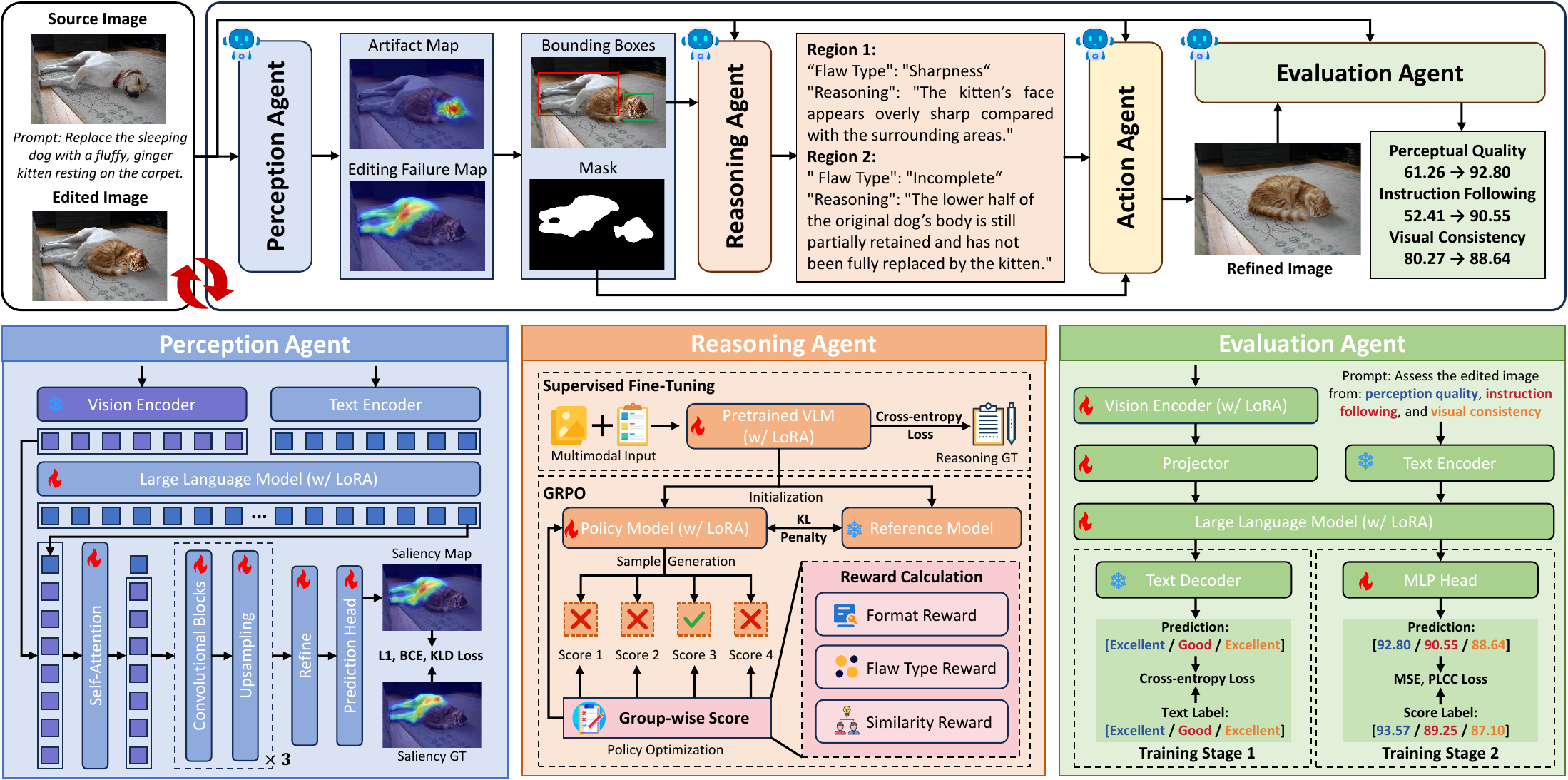}
    \caption{Overview of our EditRefiner. The framework operates as a perception-reasoning-action-evaluation loop for human-aligned post-editing correction.}
    \label{editrefiner}
\end{figure}
\subsection{Data analysis}
Through the three-stage annotation process, EditFHF-15K comprises 15,287 images, each associated with two saliency maps for artifact and editing failure. It contains 60,318 annotated artifact regions and 80,345 editing failure regions. Each annotated region is paired with a description averaging 32.6 words. Totally 687K human ratings are collected, which are aggregated to compute 45,861 MOS values across three dimensions: perceptual quality, instruction following, and visual consistency. The distribution of flaw types and MOSs are shown in Figure~\ref{editfhf}(b).
More details of annotation pipeline are provided in Section~\ref{annotation_more} of supplementary material.

\section{EditRefiner}
\subsection{Overall framework}
\textbf{EditRefiner} is a multi-agent framework for human-aligned post-editing refinement. Unlike feed-forward editing pipelines with static corrections, it operates in a closed perception-reasoning-action-evaluation loop to autonomously identify and correct flaws. By integrating perceptual cues, semantic reasoning, and evaluation, EditRefiner produces refinements better aligned with human preferences.

As illustrated in Figure~\ref{editrefiner}, the framework comprises \textbf{four collaborative agents}:
\begin{itemize}[left=12pt, labelsep=0.6em, labelwidth=0pt]
    \item \textbf{Perception Agent:} detects contextual saliency maps that highlight artifacts and editing failures in the image, providing a perceptual basis for subsequent analysis and refinement.
    \item \textbf{Reasoning Agent:} interprets perceptual cues and performs structured, human-aligned diagnostic inference to identify the flaw types and give the explanations.
    \item \textbf{Action Agent:} leverages the perception and reasoning outputs to plan and execute precise, localized re-editing operations.
    \item \textbf{Evaluation Agent:} evaluates the re-edited image against three dimensions, providing feedback that informs whether additional refinements are required.
\end{itemize}

Formally, let $I_t$ denote the edited image at iteration $t$, $I_0$ the source image, and $T$ the editing instruction. 
The \emph{Perception Agent} generates two saliency maps, $S_\mathrm{A}$ highlighting artifacts and $S_\mathrm{F}$ highlighting editing failure. Each map $S_\mathrm{i}$ is thresholded to produce a corresponding mask $M_{i,t}$ and bounding boxes $B_{i,t}$ for the flawed regions. 
\begin{equation}
M_{i,t}, B_{i,t} = \mathrm{PerceptionAgent}(\{I_0, T\}, I_t)
\end{equation}
Based on the bounding boxes $\mathrm{B}_{i,t}$, the \emph{Reasoning Agent} produces an overall description $D_{i,t}$ containing flaw types and explanations.
\begin{equation}
D_{i,t} = \mathrm{ReasoningAgent}(\{I_0, T\}, I_t, B_{i,t})
\end{equation}
The \emph{Action Agent} then applies localized corrections based on description $D_{i,t}$ and mask $M_{i,t}$ to obtain the updated image, where the task-specific descriptions and masks, corresponding to either artifacts or editing failures, are aggregated into unified representations $D_t$ and $M_t$, respectively.
\begin{equation}
I_{t+1} = \mathrm{ActionAgent}(\{I_0, T\}, I_t, \{M_{t}, D_{t}\}), \quad t \gets t+1
\end{equation}
After each re-editing step, an Evaluation Agent assesses the updated image in terms of perceptual quality ($s_v$), instruction adherence ($s_e$), and visual consistency ($s_p$), and computes the overall score ($s_{overall}$), following \cite{lmm4edit}.
\begin{equation}
s_{v}, s_{e}, s_{p} = \mathrm{EvaluationAgent}(\{I_0, T\}, I_{t+1}), \quad s_{overall}=s_{v}^{0.3}*s_{e}^{0.4}*s_{p}^{0.3}
\end{equation}
The refinement loop is terminated when there is no improvement in the overall score, or when the maximum number of editing iterations is reached. This threshold-based stopping criterion effectively prevents unnecessary refinement and avoids cascading degradation. By integrating perception, reasoning, action, and evaluation, the framework enables step-wise, interpretable, and perceptually faithful refinement with controllable iterative corrections.

\subsection{Perception Agent}
To mimic human perceptual sensitivity, we develop a perception agent to estimate a saliency map $S \in [0,1]^{H \times W}$, conditioned on both source image $I_0$, edited image $I_e$ and editing prompt $T$. This task requires jointly understanding the instruction prompt, visual content, and comparing the source and edited images, making VLMs a natural choice due to their strong multimodal fusion and visual understanding capabilities. We adopt a VLM as the backbone and attach a saliency decoder for pixel-level prediction. To address the limitation of early fusion, we concatenate the vision encoder features with the final-layer hidden states. The fused features are then fed into the saliency decoder to produce the saliency map. The model is optimized with a hybrid saliency loss $L_\text{sal}$:
\begin{equation}
L_\text{sal} = \alpha L_\text{L1}(S, \hat{S}) + \beta L_\text{BCE}(S, \hat{S}) + (1-\alpha-\beta) L_\text{KLD}(S, \hat{S}),
\end{equation}
where $\hat{S}$ is the ground-truth saliency map. The L1 loss $L_\text{L1}$ enforces pixel-level accuracy, the binary cross-entropy loss $L_\text{BCE}$ encourages the model to distinguish flawed from non-flawed regions, and the KL divergence loss $L_\text{KLD}$ aligns the predicted saliency map with human perceptual distributions. 

The predicted saliency map is thresholded with \(\tau\) to produce a binary mask \(M\) and bounding boxes \(B\).  
The bounding boxes \(B\) serve as spatial priors, providing perceptual cues to guide the analysis of Reasoning Agent, while mask \(M\) directly highlights flawed regions for Action Agent to perform localized correction.

\subsection{Reasoning Agent}
Given the bounding boxes $\{B\}$, the Reasoning Agent performs structured diagnosis, generating textual descriptions $\{D\}$ that include flaw types and reasoning statements. This requires human-aligned reasoning rather than simple classification or captioning.  

We adopt a two-stage preference alignment paradigm to train a VLM. In the first stage, \textbf{Supervised Fine-Tuning (SFT)}, the reasoning model is initialized with structured response formats, and trained using cross-entropy loss.  

In the second stage, \textbf{Group Relative Policy Optimization (GRPO)}, the model aligns its reasoning behavior with human preferences through reinforcement signals:  
\begin{equation}
L_\text{GRPO} = \mathbb{E}\Big[ \min(r_t \hat{A}_t, \text{clip}(r_t,1-\epsilon,1+\epsilon)\hat{A}_t) - \beta D_\text{KL}[\pi_\theta || \pi_\text{ref}] \Big],
\end{equation}  
where $\hat{A}_t$ captures normalized advantages computed from the reward function, which consists of format reward, flaw type reward, and similarity reward, with details provided in the supplementary. This stage ensures consistent, human-aligned reasoning across diverse flaws.

\subsection{Action Agent}
Building on the reasoning outputs, the Action Agent translates the diagnostic information \(D\) into a controllable re-editing instruction that specifies how the image should be corrected. It then provides the source image, the edited image from the previous iteration, the generated editing instruction, and the mask \(M\) highlighting the regions requiring modification as input to an editing model. The model, which supports multiple image editing operations, produces a re-edited image that progressively corrects flaws and better aligns with the desired output.

\subsection{Evaluation Agent}
The Evaluation Agent assesses the quality of re-edited images along three human-aligned dimensions: perceptual quality, instruction following, and visual consistency. It is built upon a VLM and a trainable MLP scoring head. The VLM extracts rich visual and semantic representations, which are then processed by the scoring head to predict quantitative scores. The model is trained in two stage. In the first stage, continuous scores are converted into five textual levels (bad, poor, fair, good, excellent) and optimized with cross-entropy loss. In the second stage, the model is trained using a combination of mean squared error (MSE) loss and Pearson linear correlation coefficient (PLCC) loss across all three dimensions. The losses are computed as
\begin{equation}
L_\text{score} = \sum_{d} \left( L_\text{MSE}^{(d)} + L_\text{PLCC}^{(d)} \right),
\end{equation}
where $d \in \{\text{perceptual quality, instruction following, visual consistency}\}$. This total loss ensures that the predicted scores are both accurate and well-correlated with human judgments across all three evaluation dimensions. The Evaluation Agent provides human-aligned scores to guide the loop, ensuring high-quality re-edited images and enabling appropriate termination.

The proposed multi-agent framework integrates perception-driven diagnosis, context-aware reasoning, controllable re-editing, and multi-dimensional assessing within a unified loop, enabling interpretable and autonomous refinement in better alignment with human preferences.
\begin{table}[t]
\centering
\renewcommand\arraystretch{1.2}
\caption{Performance of \textbf{EditRefiner} on EditFHF-15K, GEdit-Bench \cite{liu2025step1x-edit}, I2I-Bench \cite{wang2025i2ibenchcomprehensivebenchmarksuite}, and KRIS-Bench \cite{wu2025krisbenchbenchmarkingnextlevelintelligent}. \textbf{PQ}: Perceptual Quality, \textbf{IF}: Instruction Following, \textbf{VC}: Visual Consistency, \textbf{PP}: Physical Plausibility. Better performances over baseline are \textbf{bolded}.}
\scriptsize
   \resizebox{\linewidth}{!}{
   \begin{tabular}{l||ccc:c|cc:c|cccc:c|ccc:c}
    \toprule
      \noalign{\vspace{-1.5pt}}
     \textbf{Benchmark}&\multicolumn{4}{c}{\textbf{EditFHF-15K}}&\multicolumn{3}{c}{\textbf{GEdit-Bench \cite{liu2025step1x-edit}}}&\multicolumn{5}{c}{\textbf{I2IBench \cite{wang2025i2ibenchcomprehensivebenchmarksuite}}}&\multicolumn{4}{c}{\textbf{KRIS-Bench \cite{wu2025kris}}}\\
     \noalign{\vspace{-1.5pt}}
  \cmidrule(lr){2-5} \cmidrule(lr){6-8} \cmidrule(lr){9-13} \cmidrule(lr){14-17}
  \noalign{\vspace{-1.5pt}}
  \textbf{Method/Dimensions}&\textbf{PQ}&\textbf{IF}&\textbf{VC}&\textbf{Average}&\textbf{PQ}&\textbf{IF}&\textbf{Average}&\textbf{PQ}&\textbf{IF}&\textbf{VC}&\textbf{PP}&\textbf{Average}&\textbf{PQ}&\textbf{IF}&\textbf{VC}&\textbf{Average}\\
  \noalign{\vspace{-1.5pt}}
  \midrule
  \noalign{\vspace{-1.5pt}}
    \textbf{Qwen-Image-Edit \cite{qwenedit}}& 69.38 & 63.15 & 67.33 & 66.62 & 7.872 & 8.011 & 7.941 & 0.811 & 0.931 & 0.898 & 0.527 & 0.792 & 80.25 & 61.37 & 70.28 & 70.63 \\
    \hdashline
    (w/Reasoning Agent)& \textbf{72.06} & \textbf{65.53} & 66.24 & \textbf{67.94} & \textbf{7.946} & 7.961 & \textbf{7.954} & \textbf{0.815} & \textbf{0.933} & 0.879 & \textbf{0.543} & \textbf{0.792} & 79.06 & \textbf{63.53} & 70.24 & \textbf{70.94} \\
    $\Delta\% (\uparrow)$& +3.86 & +3.77 & -1.62 & +1.99 & +0.94 & -0.62 & +0.15 & +0.41 & +0.15 & -2.11 & +2.98 & +0.05 & -1.48 & +3.52 & -0.06 & +0.44 \\
    \hdashline
    \rowcolor{gray!10}\textbf{(w/EditRefiner)}& \textbf{74.20} & \textbf{71.50} & \textbf{72.05} & \textbf{72.58} & \textbf{8.182} & \textbf{8.431} & \textbf{8.306} & \textbf{0.836} & \textbf{0.945} & \textbf{0.920} & \textbf{0.565} & \textbf{0.816} & \textbf{82.06} & \textbf{68.53} & \textbf{72.31} & \textbf{74.30} \\
    \rowcolor{gray!10}$\Delta\%(\uparrow)$& +6.95 & +13.2 & +7.01 & +8.95 & +3.94 & +5.24 & +4.60 & +3.06 & +1.49 & +2.39 & +7.21 & +3.10 & +2.26 & +11.67 & +2.89 & +5.19 \\
    \noalign{\vspace{-1.5pt}}
    \midrule
    \noalign{\vspace{-1.5pt}}
    \textbf{NanoBanana \cite{nanobanana}}& 70.51 & 65.86 & 69.21 & 68.53 & 7.810 & 8.212 & 8.011 & 0.805 & 0.933 & 0.903 & 0.542 & 0.796 & 81.74 & 63.21 & 71.90 & 72.28 \\
    \hdashline
(w/Reasoning Agent)& \textbf{70.96} & \textbf{68.28} & \textbf{71.55} & \textbf{70.26} & \textbf{7.894} & \textbf{8.285} & \textbf{8.089} & 0.803 & \textbf{0.938} & \textbf{0.904} & \textbf{0.557} & \textbf{0.800} & \textbf{82.63} & \textbf{64.08} & \textbf{72.01} & \textbf{72.91} \\
    $\Delta\%(\uparrow)$& +0.64 & +3.68 & +3.38 & +2.57 & +1.08 & +0.89 & +0.98 & -0.31 & +0.53 & +0.19 & +2.71 & +0.59 & +1.09 & +1.38 & +0.15 & +0.87 \\
    \hdashline
    \rowcolor{gray!10}\textbf{(w/EditRefiner)}& \textbf{75.20} & \textbf{70.45} & \textbf{73.82} &\textbf{73.16} & \textbf{8.216} & \textbf{8.644} & \textbf{8.430} & \textbf{0.830} & \textbf{0.938} & \textbf{0.922} & \textbf{0.572} & \textbf{0.816} & \textbf{84.17} & \textbf{68.27} & \textbf{75.82} & \textbf{76.09} \\
    \rowcolor{gray!10}$\Delta\%(\uparrow)$ & +6.65 & +6.97 & +6.66 & +6.76 & +5.20 & +5.26 & +5.23 & +3.03 & +0.58 & +2.21 & +5.48 & +2.50 & +2.97 & +8.00 & +5.45 & +5.48 \\
    \noalign{\vspace{-1.5pt}}
    \midrule
    \noalign{\vspace{-1.5pt}}
    \textbf{Seedream4.0 \cite{seedream4}}& 68.27 & 66.18 & 65.24 & 66.56 & 7.650 & 7.941 & 7.795 & 0.815 & 0.917 & 0.891 & 0.519 & 0.786 & 82.56 & 62.49 & 69.88 & 71.64 \\
    \hdashline
(w/Reasoning Agent)& \textbf{70.04} & 65.12 & 65.15 & \textbf{66.77} & \textbf{7.722} & \textbf{7.965} & \textbf{7.843} & \textbf{0.822} & \textbf{0.920} & 0.883 & \textbf{0.526} & \textbf{0.788} & 82.04 & \textbf{64.12} & \textbf{70.25} & \textbf{72.14} \\
    $\Delta\%(\uparrow)$& +2.59 & -1.60 & -0.14 & 0.32 & +0.94 & +0.30 & +0.62 & +0.87 & +0.32 & -0.95 & +1.39 & +0.28 & -0.63 & +2.61 & +0.53 & +0.84 \\
     \hdashline
    \rowcolor{gray!10}\textbf{(w/EditRefiner)}& \textbf{71.46} & \textbf{70.28} & \textbf{70.84} & \textbf{70.86} & \textbf{7.981} & \textbf{8.244} & \textbf{8.113} & \textbf{0.838} & \textbf{0.937} & \textbf{0.906} & \textbf{0.539} & \textbf{0.805} & \textbf{85.28} & \textbf{66.15} & \textbf{73.04} & \textbf{74.82} \\
    \rowcolor{gray!10}$\Delta\%(\uparrow)$ & +4.67 & +6.20 & +8.58 & +6.48 & +4.33 & +3.82 & +4.07 & +2.80 & +2.19 & +1.68 & +3.74 & +2.46 & +3.30 & +5.86 & +4.52 & +4.56 \\
    \noalign{\vspace{-1.5pt}}
    \bottomrule
  \end{tabular}}
  \label{overall}
\end{table}
\begin{figure}
    \centering
    \includegraphics[width=1\linewidth]{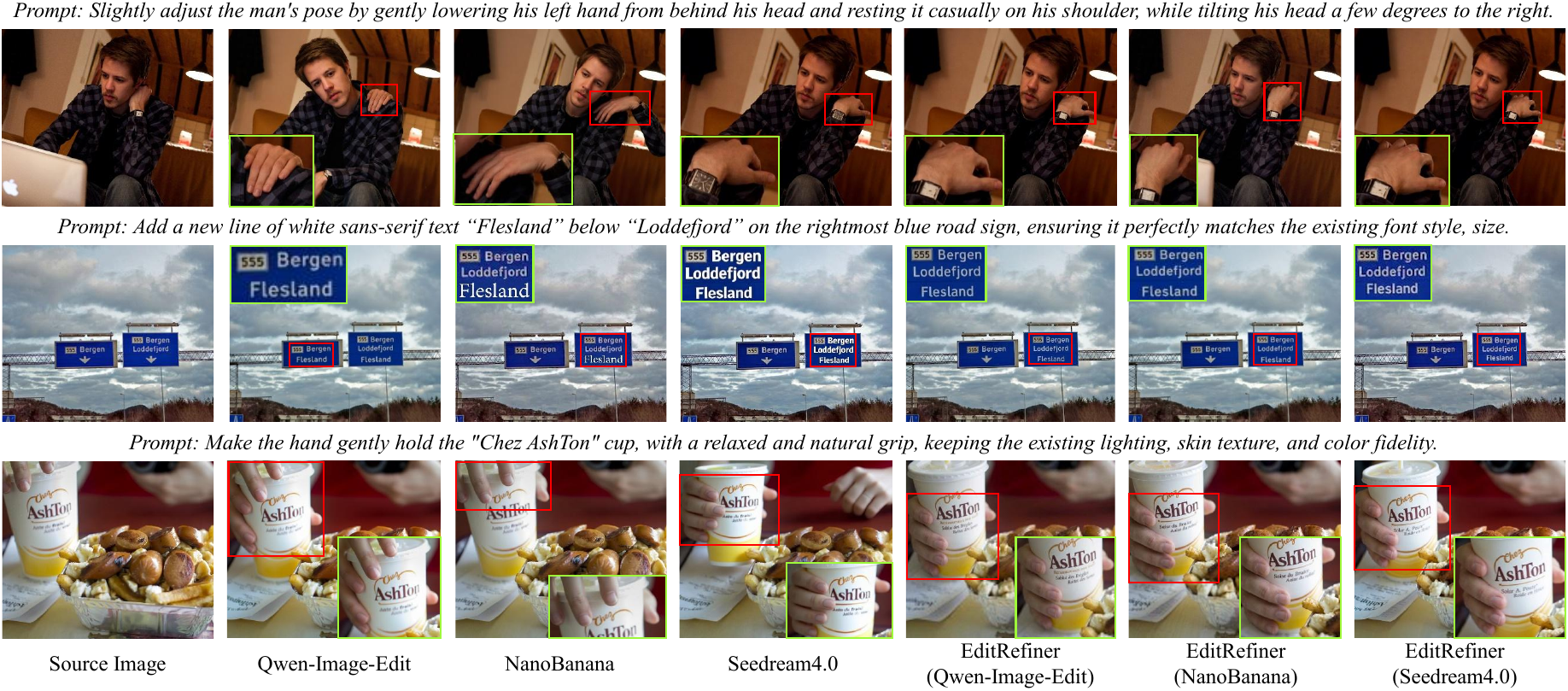}
    \caption{Example results from advanced TIE models and results with our EditRefiner. Additional results are provided in the supplementary material.}
    \label{examples}
\end{figure}

\begin{table}[t]
  \centering
  \caption{Quantitative evaluation and ablation of the Perception Agent on flaw-aware saliency prediction. $\spadesuit$ hand-crafted methods, $\heartsuit$ deep learning-based methods, \ding{72} general-purpose VLMs. The fine-tuned results are marked with \raisebox{0.5ex}{\scriptsize \ding{91}}. 
The best results are highlighted in \mredbf{red}, and the second-best results are highlighted in \mbluebf{blue}.}
  \renewcommand\arraystretch{1.1}  
  \resizebox{1\textwidth}{!}{  
  \setlength{\tabcolsep}{16pt}
    \begin{tabular}{l||ccccc|ccccc}  
      \toprule
        \noalign{\vspace{-1.5pt}}
      \textbf{Dimension}&\multicolumn{5}{c}{\textbf{Artifact Region}}&\multicolumn{5}{c}{\textbf{Editing Failure Region}}\\
        \noalign{\vspace{-1.5pt}}
  \cmidrule(lr){2-6} \cmidrule(lr){7-11}
    \noalign{\vspace{-1.5pt}}
      \textbf{Method/Metric} & \textbf{AUC-Judd} $\uparrow$ & \textbf{ NSS}$\uparrow$ & \textbf{ CC} $\uparrow$& \textbf{ SIM}$\uparrow$ &\textbf{ KLD}$\downarrow$& \textbf{AUC-Judd} $\uparrow$ & \textbf{ NSS}$\uparrow$ & \textbf{ CC} $\uparrow$& \textbf{ SIM}$\uparrow$ &\textbf{ KLD}$\downarrow$ \\
        \noalign{\vspace{-1.5pt}}
      \midrule
        \noalign{\vspace{-1.5pt}}
      $\spadesuit$AIM~\cite{NIPS2005_0738069b} & 0.7223 & 1.1476 & 0.1689 & 0.0794 & 3.0217
& 0.6395 & 0.5283 & 0.1427 & 0.0416 & 7.3412 \\
      $\spadesuit$SMVJ~\cite{NIPS2007_708f3cf8} & 0.7169 & 0.7093 & 0.0786 & 0.0631 & 3.3698
& 0.5821 & 0.3538 & 0.0714 & 0.0426 & 7.5923 \\
      $\spadesuit$SWD~\cite{zhao2011visual} & 0.7176 & 0.5294 & 0.0708 & 0.0757 & 3.5219
& 0.5912 & 0.3876 & 0.0941 & 0.0493 & 6.8115 \\
      $\spadesuit$CA~\cite{5539929} & 0.5528 & 0.4619 & 0.1016 & 0.0687 & 3.3325
& 0.5234 & 0.0941 & 0.0548 & 0.0615 & 7.6489 \\
      \hdashline
      $\heartsuit$ResNet-50 \cite{resnet}\raisebox{0.5ex}{\scriptsize \ding{91}}& 0.7816 & 1.4498 & 0.4674 & 0.1768 & 3.0173
& 0.6882 & 0.8536 & 0.1492 & 0.0731 & 4.1964 \\
      $\heartsuit$SALICON~\cite{Huang_2015_ICCV}\raisebox{0.5ex}{\scriptsize \ding{91}} & 0.8227 & 1.5791 & 0.5028 & 0.2721 & 2.7194
& 0.7093 & 0.9845 & 0.4586 & 0.2487 & 3.5635 \\
      $\heartsuit$MLNet~\cite{mlnet2016}\raisebox{0.5ex}{\scriptsize \ding{91}} & 0.7536 & 1.5462 & 0.3527 & 0.2394 & 2.2371
& 0.7235 & 0.9588 & 0.3186 & 0.2243 & 3.4927 \\

      
      $\heartsuit$SAM-ResNet~\cite{cornia2018predicting}\raisebox{0.5ex}{\scriptsize \ding{91}} & 0.8357 & 1.8574 & 0.4031 & 0.4469 & \mbluebf{2.0716}
& 0.7416 & 0.9821 & 0.3724 & 0.3015 & 2.8478 \\

      $\heartsuit$TranSalNet~\cite{Lou2022TranSalNetTP}\raisebox{0.5ex}{\scriptsize \ding{91}} & 0.8034 & 1.7481 & 0.4598 & 0.4013 & 2.8692
& 0.7438 & 1.4426 & 0.4187 & 0.3045 & 3.1076 \\

      $\heartsuit$RichHF~\cite{liang2024rich}\raisebox{0.5ex}{\scriptsize \ding{91}} & \mbluebf{0.8508} & 1.7976 & 0.4739 & 0.4298 & 2.6721
& 0.7523 & \mbluebf{1.6247} & 0.4315 & 0.3326 & 2.8914 \\
      \hdashline
      \ding{72}Ovis2.5-9B~\cite{ovis25} & 0.6137 & 0.4184 & 0.1693 & 0.2472 & 7.4321
& 0.6598 & 0.6845 & 0.1487 & 0.2319 & 6.8123 \\

      \ding{72}InternVL3.5-8B~\cite{internvl3_5}  & 0.6046 & 0.7713 & 0.2096 & 0.3087 & 6.9341
& 0.6732 & 0.9024 & 0.1683 & 0.2896 & 5.1875 \\
      \ding{72}Qwen3-VL-8B~\cite{qwen3} & 0.6121 & 0.4215 & 0.1704 & 0.2496 & 6.4387
& 0.6814 & 1.0926 & 0.1511 & 0.3042 & 4.8956 \\

    \ding{72}ChatGPT-5 \cite{openai2025chatgpt5}& 0.7463 & 0.7217 & 0.2612 & 0.2913 & 6.0064
& 0.7138 & 1.2034 & 0.2526 & 0.3035 & 3.5872 \\

        \ding{72}Gemini-3.1-Pro \cite{gemini3} & 0.7071 & 0.7198 & 0.2597 & 0.2896 & 6.0142
& 0.7316 & 1.3115 & 0.3013 & 0.3224 & 3.3619 \\

        \hdashline
              \rowcolor{gray!20} 
      Ours (Ovis2.5-9B)\raisebox{0.5ex}{\scriptsize \ding{91}} & 0.8035 & 1.5086 & 0.4184 & 0.3795 & 2.2042
& 0.7794 & 1.5267 & 0.3421 & 0.3618 & 2.9937 \\

      \rowcolor{gray!20} 
      Ours (InternVL3.5-8B)\raisebox{0.5ex}{\scriptsize \ding{91}} & 0.8431 & \mbluebf{1.8476} & \mbluebf{0.4792} & \mbluebf{0.4687} & 2.2073
& \mbluebf{0.8015} & 1.6014 & \mbluebf{0.4528} & \mbluebf{0.4212} & \mbluebf{2.3968} \\
    \rowcolor{gray!20} 
      \textbf{Ours (Qwen3-VL-8B)}\raisebox{0.5ex}{\scriptsize \ding{91}} & \mredbf{0.8720} & \mredbf{2.0256} & \mredbf{0.5550} & \mredbf{0.4901}& \mredbf{1.9515} & \mredbf{0.8306} & \mredbf{1.8925} & \mredbf{0.5142} &\mredbf{0.4578} & \mredbf{2.2191} \\
        \noalign{\vspace{-1.5pt}}
      \bottomrule
    \end{tabular}
  }
  \label{perception}
\end{table}

\begin{figure}[t]
    \centering
    \includegraphics[width=1\linewidth]{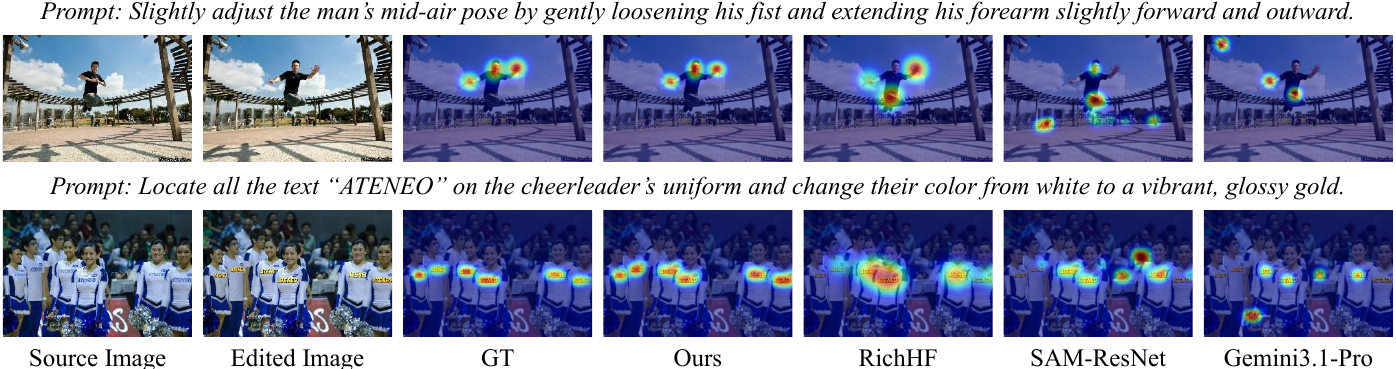}
    \caption{Visualization of saliency map prediction. Our method produces sharper and more precise localization. The first row highlights artifact regions, while the second row shows editing failure regions.}
    \label{region}
\end{figure}

\begin{table}[t]
  \centering
  \caption{Quantitative evaluation and ablation of the Reasoning Agent. The best results are \textbf{bolded}.}
  \renewcommand\arraystretch{1.1}  
  \resizebox{1\textwidth}{!}{  
    \setlength{\tabcolsep}{10pt} 
    \begin{tabular}{l||ccccc|ccccc}  
      \toprule
        \noalign{\vspace{-1.5pt}}
      \normalsize
      \textbf{Dimension}&\multicolumn{5}{c}{\textbf{Artifact Region}}&\multicolumn{5}{c}{\textbf{Editing Failure Region}}\\
        \noalign{\vspace{-1.5pt}}
  \cmidrule(lr){2-6} \cmidrule(lr){7-11}
    \noalign{\vspace{-1.5pt}}
      \textbf{Method/Metric} & \textbf{Accuracy} $\uparrow$ & \textbf{ SimCSE}$\uparrow$ & \textbf{Word2Vec}$\uparrow$ & \textbf{Meteor} $\uparrow$ & \textbf{ROUGE} $\uparrow$& \textbf{Accuracy} $\uparrow$ & \textbf{ SimCSE}$\uparrow$ & \textbf{Word2Vec}$\uparrow$ & \textbf{Meteor} $\uparrow$ & \textbf{ROUGE} $\uparrow$ \\
        \noalign{\vspace{-1.5pt}}
      \midrule
        \noalign{\vspace{-1.5pt}}
      ChatGPT-5 \cite{openai2025chatgpt5} & 62.18\% &0.6728& 0.5846 & 0.1550 & 0.1304&68.25\%&0.6532&0.5429&0.1526&0.1280 \\
Gemini-3.1-Pro \cite{gemini3} & 60.91\% & 0.6883 & 0.6281 & 0.1689 & 0.1137 
& 66.48\% & 0.6637 & 0.5824 & 0.1671 & 0.1112 \\
\hdashline
Ovis2.5-9B~\cite{ovis25} & 58.12\% & 0.6694 & 0.6136 & 0.1665 & 0.0746 
& 63.27\% & 0.6468 & 0.5681 & 0.1642 & 0.0729 \\
Ovis2.5-9B + GRPO & 64.36\% & 0.7057 & 0.6628 & 0.2226 & 0.1518 
& 69.01\% & 0.6812 & 0.6153 & 0.1704 & 0.1295 \\
Ovis2.5-9B + SFT & 72.96\% & 0.8012 & 0.7495 & 0.3586 & 0.3389 
& 80.11\% & 0.8036 & 0.7132 & 0.3508 & 0.3165 \\
\rowcolor{gray!20}
\textbf{Ours (GRPO + SFT)} & \textbf{80.67\%} & \textbf{0.8461} & \textbf{0.7819} & \textbf{0.4052}& \textbf{0.3554} 
& \textbf{82.93\%} & \textbf{0.8264} & \textbf{0.7376} & \textbf{0.3967}& \textbf{0.3428} \\
\hdashline
InternVL3.5-8B~\cite{internvl3_5} & 58.73\% & 0.6759 & 0.6214 & 0.1687 & 0.0979 
& 64.05\% & 0.6526 & 0.5741 & 0.1659 & 0.0956 \\
InternVL3.5-8B + GRPO & 60.82\% & 0.7216 & 0.6278 & 0.1815 & 0.1243 
& 66.14\% & 0.6968 & 0.5856 & 0.1782 & 0.1217 \\
InternVL3.5-8B + SFT & 77.88\% & 0.8389 & 0.7762 & 0.3948 & 0.3561
& 79.22\% & 0.8197 & 0.7318 & 0.3875 & 0.3439\\
\rowcolor{gray!20}
\textbf{Ours (GRPO + SFT)} & \textbf{79.94\%} & \textbf{0.8443} & \textbf{0.7846} & \textbf{0.4206}& \textbf{0.3789} 
& \textbf{81.02\%} & \textbf{0.8258} & \textbf{0.7397} & \textbf{0.4093}& \textbf{0.3642} \\
\hdashline
Qwen3-VL-8B~\cite{qwen3} & 57.38\% & 0.6835 & 0.6082 & 0.1661 & 0.1049 
& 62.77\% & 0.6613 & 0.5667 & 0.1634 & 0.1023 \\
Qwen3-VL-8B + GRPO & 70.14\% & 0.7498 & 0.7084 & 0.2183 & 0.2482 
& 72.63\% & 0.7451 & 0.6529 & 0.2451 & 0.2036 \\
Qwen3-VL-8B + SFT & 79.31\% & 0.8324 & 0.7649 & 0.3567 & 0.3268 
& 80.58\% & 0.8127 & 0.7196 & 0.3489 & 0.3149 \\
\rowcolor{gray!20}
\textbf{Ours (GRPO + SFT)} & \textbf{81.08\%} & \textbf{0.8527} & \textbf{0.7856} & \textbf{0.3942}& \textbf{0.3704} 
& \textbf{84.36\%} & \textbf{0.8338} & \textbf{0.7421} & \textbf{0.3859}& \textbf{0.3485}  \\
  \noalign{\vspace{-1.5pt}}
      \bottomrule
    \end{tabular}
  }
  \label{reasoning}
\end{table}

\begin{table*}[t]

\caption{
Quantitative evaluation of the Evaluation Agent.
$\spadesuit$ Hand-crafted VQA metrics, $\clubsuit$ deep learning-based VQA methods, \ding{72} general-purpose VLMs, \ding{73} editing-oriented VLMs. 
The fine-tuned results are marked with \raisebox{0.5ex}{\scriptsize \ding{91}}. 
The best results are highlighted in \mredbf{red}, and the second-best results are highlighted in \mbluebf{blue}.
}
\centering
\renewcommand\arraystretch{1.2}  
    \setlength{\tabcolsep}{13pt}
\resizebox{1\textwidth}{!}{\begin{tabular}{l||ccc|ccc|ccc|ccc}
\toprule
  \noalign{\vspace{-1.5pt}}
\textbf{Dimension}& \multicolumn{3}{c}{\textbf{Perceptual Quality}} & \multicolumn{3}{c}{\textbf{Instruction Following}} & \multicolumn{3}{c}{\textbf{Visual Consistency}}& \multicolumn{3}{c}{\textbf{Overall Score}} \\
  \noalign{\vspace{-1.5pt}}
\cmidrule(lr){2-4}
\cmidrule(lr){5-7}
\cmidrule(lr){8-10}
\cmidrule(lr){11-13}
  \noalign{\vspace{-1.5pt}}
\textbf{Methods/Metrics} &\textbf{SRCC}&\textbf{KRCC}&\textbf{PLCC}&\textbf{SRCC}&\textbf{KRCC}&\textbf{PLCC}&\textbf{SRCC}&\textbf{KRCC}&\textbf{PLCC}&\textbf{SRCC}&\textbf{KRCC}&\textbf{PLCC} \\
\hline
$\spadesuit$GMSD \cite{GMSD} & 0.1274 & 0.0776 & 0.1853 & 0.1081 & 0.0723 & 0.1059 & 0.1068 & 0.1035 & 0.1534& 0.1641 & 0.0988 & 0.1628 \\
$\spadesuit$SSIM \cite{SSIM} & 0.1004 & 0.0907 & 0.1199 & 0.1055 & 0.0698 & 0.1080 & 0.1165 & 0.0113 & 0.1103 & 0.1470 & 0.1023 & 0.1602\\
$\spadesuit$VIF \cite{VIF} & 0.2196 & 0.1786 & 0.2761 & 0.0959 & 0.0640 & 0.1035 & 0.2007 & 0.1657 & 0.2650 & 0.2121 & 0.1801 & 0.2326 \\
\hdashline
$\clubsuit$LPIPS \cite{LPIPS} & 0.3435 & 0.1960 & 0.3032 & 0.1244 & 0.0823 & 0.3332 & 0.2457 & 0.1976 & 0.3083 & 0.3079 & 0.2486 & 0.3249 \\
$\clubsuit$CVRKD\raisebox{0.5ex}{\scriptsize \ding{91}} \cite{CVRKD} & 0.6507 & 0.4721 & 0.7244 & 0.1802 & 0.1209 & 0.1963 & 0.5371 & 0.3814 & 0.6208 & 0.4507 & 0.3221 & 0.4622 \\
$\clubsuit$AHIQ\raisebox{0.5ex}{\scriptsize \ding{91}} \cite{AHIQ} & 0.7172 & 0.5321 & \mbluebf{0.7779} & 0.2385 & 0.1612 & 0.2510 & 0.6583 & 0.4136 & 0.6775 & 0.4880 & 0.3209 & 0.5015 \\
$\clubsuit$Q-Align\raisebox{0.5ex}{\scriptsize \ding{91}} \cite{qalign} & \mbluebf{0.7565} & \mbluebf{0.6640} & 0.7650 & 0.5663 & 0.4120 & 0.5971 & 0.6856 & 0.4948 & 0.7123 & 0.6095 & 0.4230 & 0.6208 \\
\hdashline
\ding{72}Ovis2.5-9B \cite{ovis25} & 0.1010 & 0.1008 & 0.1726 & 0.1121 & 0.0886 & 0.1428 & 0.2209 & 0.1897 & 0.1345 & 0.2017 & 0.1661 & 0.2405 \\
\ding{72}Qwen3-VL-8B \cite{qwen3}& 0.2455 & 0.2075 & 0.2236 & 0.2412 & 0.2327 & 0.2365 & 0.2772 & 0.2604 & 0.3292 & 0.2536 & 0.2315 & 0.2601 \\
\ding{72}InternVL3-8B \cite{internvl3_5}& 0.2328 & 0.2079 & 0.2747 & 0.2059 & 0.1859 & 0.1649 & 0.3226 & 0.2796 & 0.3618 & 0.2531 & 0.2242 & 0.2621 \\
\ding{72}ChatGPT-5 \cite{gpt5}& 0.3309 & 0.2631 & 0.3749 & 0.2786 & 0.2127 & 0.3133 & 0.3383 & 0.2343 & 0.4156 & 0.3559 & 0.2961 & 0.3680 \\
\ding{72}Gemini-3.1-Pro \cite{gemini3} & 0.3115 & 0.2874 & 0.3386 & 0.2927 & 0.2618 & 0.2966 & 0.3009 & 0.2396 & 0.3704 & 0.3410 & 0.2927 & 0.3651 \\
\hdashline
\ding{73}LMM4Edit \cite{lmm4edit} & 0.3509 & 0.2424 & 0.3651 & 0.3385 & 0.2109 & 0.3457 & 0.3555 & 0.2145 & 0.3552 & 0.3883 & 0.3006 & 0.3851 \\
\ding{73}EditScore (Qwen3) \cite{editscore} & 0.2068 & 0.1515 & 0.1842 & 0.2632 & 0.2262 & 0.1316 & 0.2886 & 0.2451 & 0.1274 & 0.2829 & 0.2270 & 0.2497 \\
\ding{73}EditReward (MiMo) \cite{editreward}& 0.3276 & 0.2519 & 0.3357 & 0.0508 & 0.0336 & 0.0683 & 0.2627 & 0.2087 & 0.2679 & 0.3037 & 0.2647 & 0.3251 \\
\ding{73}EditHF \cite{xu2026edithf1mmillionscalerichhuman} & 0.2082 & 0.1427 & 0.1416 & 0.1325 & 0.0886 & 0.1191 & 0.1520 & 0.1023 & 0.1365 & 0.1412 & 0.1225 & 0.1323 \\
\hline
  \rowcolor{gray!20}
  Ours (Ovis2.5-9B)\raisebox{0.5ex}{\scriptsize \ding{91}} & 0.7233 & 0.5371 & 0.7471 & 0.7100 & 0.5281 & 0.7212 & 0.6797 & 0.4126 & 0.7295 & 0.7643 & 0.6626 & 0.7826 \\
  \rowcolor{gray!20}
  Ours (InternVL3.5-8B)\raisebox{0.5ex}{\scriptsize \ding{91}}& 0.7335 & 0.5044 & 0.7278 & \mbluebf{0.7244} & \mbluebf{0.5409} & \mbluebf{0.7353} & \mbluebf{0.6952} & \mbluebf{0.5088} & \mbluebf{0.7332} & \mbluebf{0.8177} & \mbluebf{0.7221} & \mbluebf{0.8346} \\
  \rowcolor{gray!20}  
\textbf{Ours (Qwen3-VL-8B)}\raisebox{0.5ex}{\scriptsize \ding{91}}& \mredbf{0.7852} & \mredbf{0.6686} & \mredbf{0.8014} & \mredbf{0.7522} & \mredbf{0.6508} & \mredbf{0.7641} & \mredbf{0.7885} & \mredbf{0.6412} & \mredbf{0.7969} & \mredbf{0.8723} & \mredbf{0.7235} & \mredbf{0.8842} \\
\noalign{\vspace{-1.5pt}}
\bottomrule
\end{tabular}}
\label{evaluation}
\end{table*}

\begin{table*}[!htbp]

\caption{
Ablation of turn number. The best results are \textbf{bolded}.
}
\centering
\renewcommand\arraystretch{1.1}  
    \setlength{\tabcolsep}{9pt}
\resizebox{1\textwidth}{!}{\begin{tabular}{l||ccc|ccc|ccc|ccc}
\toprule
  \noalign{\vspace{-1.5pt}}
Benchmark& \multicolumn{3}{c}{EditHF-15K} & \multicolumn{3}{c}{GEdit-Bench \cite{liu2025step1x-edit}} & \multicolumn{3}{c}{I2IBench \cite{wang2025i2ibenchcomprehensivebenchmarksuite}}&\multicolumn{3}{c}{KRIS-Bench \cite{wu2025kris}} \\
  \noalign{\vspace{-1.5pt}}
\cmidrule(lr){2-4}
\cmidrule(lr){5-7}
\cmidrule(lr){8-10}
\cmidrule(lr){11-13}
  \noalign{\vspace{-1.5pt}}
Turns/Method&Qwen \cite{qwenedit}&Nano \cite{nanobanana}&Seed \cite{seedream4}&Qwen \cite{qwenedit}&Nano \cite{nanobanana}&Seed \cite{seedream4}&Qwen \cite{qwenedit}&Nano \cite{nanobanana}&Seed \cite{seedream4}&Qwen \cite{qwenedit}&Nano \cite{nanobanana}&Seed \cite{seedream4} \\ 
  \noalign{\vspace{-1.5pt}}
\midrule
  \noalign{\vspace{-1.5pt}}
Turn 0&66.62&68.53&66.56&7.941&8.011&7.795&0.792&0.796&0.786&70.63&72.28&71.64\\
Turn 1&69.28&71.02&68.91&8.192&8.244&8.036&0.807&0.809&0.800&72.74&74.51&73.33\\
Turn 2&71.11&71.36&69.84&8.167&8.298&8.052&0.805&0.808&0.801&72.98&74.83&73.58\\
Turn 3&72.52&72.94&70.40&8.233&8.421&8.101&0.809&0.811&0.803&73.26&75.21&73.92\\
Turn 4&\textbf{72.64}&73.05&70.72&\textbf{8.311}&\textbf{8.438}&8.097&0.814&0.814&0.801&\textbf{74.34}&76.08&\textbf{74.89}\\
  \rowcolor{gray!20}
Auto Stop&72.58&\textbf{73.16}&\textbf{70.86}&8.306&8.430&\textbf{8.113}&\textbf{0.816}&\textbf{0.816}&\textbf{0.805}&74.30&\textbf{76.09}&74.82\\
  \noalign{\vspace{-1.5pt}}
\bottomrule
\end{tabular}}
\label{turnablation}
\end{table*}
\section{Experiments}

\subsection{Experimental setup}
\textbf{Training Configuration.}
We train the EditRefiner on our EditFHF-15K, which is split to training, validating and testing set with a 4:1:1 ratio. The Perception, Reasoning, and Evaluation Agents are fine-tuned via LoRA \cite{lora} based on the Qwen3-VL-8B \cite{qwen3} backbone. 
Inference is fully automatic, converging in an average of 2.53 iterations with a maximum of 4.

\textbf{Evaluation Datasets and Metrics.}
To evaluate the refinement capability and validate generalization, we report results on EditFHF-15K, GEdit-Bench \cite{liu2025step1x-edit}, I2I-Bench \cite{wang2025i2ibenchcomprehensivebenchmarksuite}, and KRIS-Bench \cite{wu2025krisbenchbenchmarkingnextlevelintelligent}. Results are averaged over five runs with different random seeds for stability. We use the Evaluation Agent of EditRefiner as the evaluation metric on EditFHF-15K, while following the respective evaluation protocols for the other benchmarks. For individual modules, following \cite{8315047,shen2026agentic}, Perceptual Agent is evaluated using CC, SIM, KLD, AUC-Judd, and NSS. Reasoning Agent is assessed via distortion classification accuracy and semantic alignment with ROUGE~\cite{lin-2004-rouge}, METEOR~\cite{banerjee-lavie-2005-meteor}, Word2Vec~\cite{mikolov2013efficientestimationwordrepresentations}, and SimCSE~\cite{gao2021simcse}. Evaluation Agent is measured using SRCC, KRCC, and PLCC.

\subsection{Comparison sesults}
\subsubsection{Overall system performance}
As shown in Table~\ref{overall}, the performance of EditRefiner across various benchmarks demonstrates its strong capability in enhancing image editing. The Reasoning Agent alone can provide performance gains, but its effectiveness is limited due to the lack of explicit local region information. In contrast, EditRefiner consistently achieves substantial improvements across different evaluation settings. Figure~\ref{examples} further presents qualitative comparisons across various scenes and prompt conditions.

\subsubsection{Individual agent analysis}
\textbf{Perception Agent.} Table~\ref{perception} compares our Perception Agent with conventional saliency detectors, learning-based methods, and general-purpose VLMs. Hand-crafted methods rely on low-level cues and fail to capture semantics, while learning-based approaches lack editing-specific understanding, leading to inaccurate localization. General-purpose VLMs also perform poorly due to insufficient task grounding. In contrast, our method achieves state-of-the-art performance across all metrics. Figure~\ref{region} further shows that our saliency maps exhibit sharper spatial focus and better alignment with human annotations, providing a reliable basis for subsequent reasoning and refinement.

\textbf{Reasoning Agent.} Table~\ref{reasoning} presents results under different training strategies. Across VLM backbones, our method consistently improves all metrics, showing that joint SFT and GRPO training enhances reasoning accuracy and better aligns with human preferences than either alone. After training on our dataset, the Reasoning Agent surpasses advanced closed-source models such as ChatGPT-5 and Gemini-3.1-Pro, demonstrating strong capability in identifying editing-related reasoning gaps and enabling human-aligned adaptation.

\textbf{Evaluation Agent.} Table~\ref{evaluation} compares the Evaluation Agent with existing VQA methods. Hand-crafted methods perform poorly due to focus on low-level distortions, while deep learning-based approaches improve visual assessment but remain limited in evaluating editing alignment. General-purpose VLMs lack task-specific understanding, and editing-oriented VLMs still struggle with accurate scoring for high-quality results. In contrast, our Evaluation Agent achieves the best alignment with human preferences across all dimensions, providing reliable feedback for iterative refinement.

\subsection{Ablation study}
Table~\ref{perception}, \ref{reasoning}, \ref{evaluation}, show the performance of the agents across different VLM backbones, where consistent improvements are observed, validating the effectiveness of our training strategy and our dataset. 
We further conduct ablations on the number of refinement turns, as shown in Table~\ref{turnablation}, indicating significant improvements within the first 1-2 iterations, while performance gradually saturates at 3-4 turns. Notably, the 4th iteration does not consistently surpass our auto-stop strategy, as excessive re-editing not only increases computational cost but also leads to error accumulation and potential subject drift.
Additional ablation studies are provided in Section~\ref{more_ablation} of the supplementary material.

\section{Conclusion}
In this paper, we introduce EditFHF-15K, a large-scale dataset of fine-grained human feedback that provides rich annotations of editing flaw regions, textual reasoning, and human ratings across diverse editing models and tasks. Built upon this dataset, we propose EditRefiner, a hierarchical and interpretable agentic framework that reformulates image refinement as a perception-reasoning-action-evaluation loop. Experiments show that EditRefiner outperforms state-of-the-art methods in flaw localization, reasoning, and human preference alignment, highlighting its potential for self-corrective, human-aligned image editing.

\newpage
\clearpage
\bibliographystyle{splncs04}
\bibliography{main}

\end{document}